\documentclass[5p]{elsarticle}

\usepackage{lineno,hyperref}
\usepackage{amsfonts}
\usepackage{amsmath}
\usepackage{amssymb}
\usepackage{mathrsfs}
\usepackage{booktabs}
\usepackage{multirow}
\usepackage{graphicx}
\usepackage{bm}
\usepackage{caption}

\usepackage[titlenumbered,ruled]{algorithm2e} 
\DeclareMathOperator*{\argmax}{arg\,max}

\journal{Journal of \LaTeX\ Templates}

\begin{document}

\begin{frontmatter}

\title{A Targeted Universal Attack on Graph Convolutional Network}

\author{Jiazhu Dai\corref{mycorrespondingauthor}}
\cortext[mycorrespondingauthor]{Corresponding author.}
\ead{daijz@shu.edu.cn}
\author{Weifeng Zhu}
\author{Xiangfeng Luo}
\address{School of Computer Engineering and Science, Shanghai University, China}


\begin{abstract}
Graph-structured data exist in numerous applications in real life. As a state-of-the-art graph neural network, the graph convolutional network (GCN) plays an important role in processing graph-structured data. However, a recent study reported that GCNs are also vulnerable to adversarial attacks, which means that GCN models may suffer malicious attacks with unnoticeable modifications of the data. Among all the adversarial attacks on GCNs, there is a special kind of attack method called the universal adversarial attack, which generates a perturbation that can be applied to any sample and causes GCN models to output incorrect results. Although universal adversarial attacks in computer vision have been extensively researched, there are few research works on universal adversarial attacks on graph structured data. In this paper, we propose a targeted universal adversarial attack against GCNs. Our method employs a few nodes as the attack nodes. The attack capability of the attack nodes is enhanced through a small number of fake nodes connected to them. During an attack, any victim node will be misclassified by the GCN as the attack node class as long as it is linked to them. The experiments on three popular datasets show that the average attack success rate of the proposed attack on any victim node in the graph reaches 83\% when using only 3 attack nodes and 6 fake nodes. We hope that our work will make the community aware of the threat of this type of attack and raise the attention given to its future defense.
\end{abstract}


\end{frontmatter}


\section{Introduction}

Graph structured data have been widely used to model real-world systems for centuries. In recent years, graph neural networks have played an important role in the field of analysis and prediction using graph structured data. Graph convolutional networks (GCNs)\cite{7_semi_conv,14_SGC}, as an extension of convolutional neural networks, are one of the most popular models for processing graph structured data, e.g., large social networks, biological science (cell interaction), physics simulations and knowledge graphs.

Despite various applications of GCNs, they are still vulnerable to adversarial attacks, that is, even slight perturbations in a graph can lead to GCN model misclassification. For example, the GCN model could be easily fooled to misclassify a specific node or a group of nodes in a graph through perturbations such as modifications in graph topologies or the attributes of nodes (Dai et al.\cite{2_Dai}). Of all the attack methods against GCNs, the universal adversarial attack, where the GCN model is deceived to misclassify any node as classes other than their ground truth class using the same perturbation, is a new threat to GCNs. However, existing research on universal adversarial attacks mainly focuses on computer vision tasks with convolutional neural networks, and few efforts have been made to study the threat of universal adversarial attacks to GCNs.

\begin{figure}
    \centering
    \includegraphics[scale=0.17]{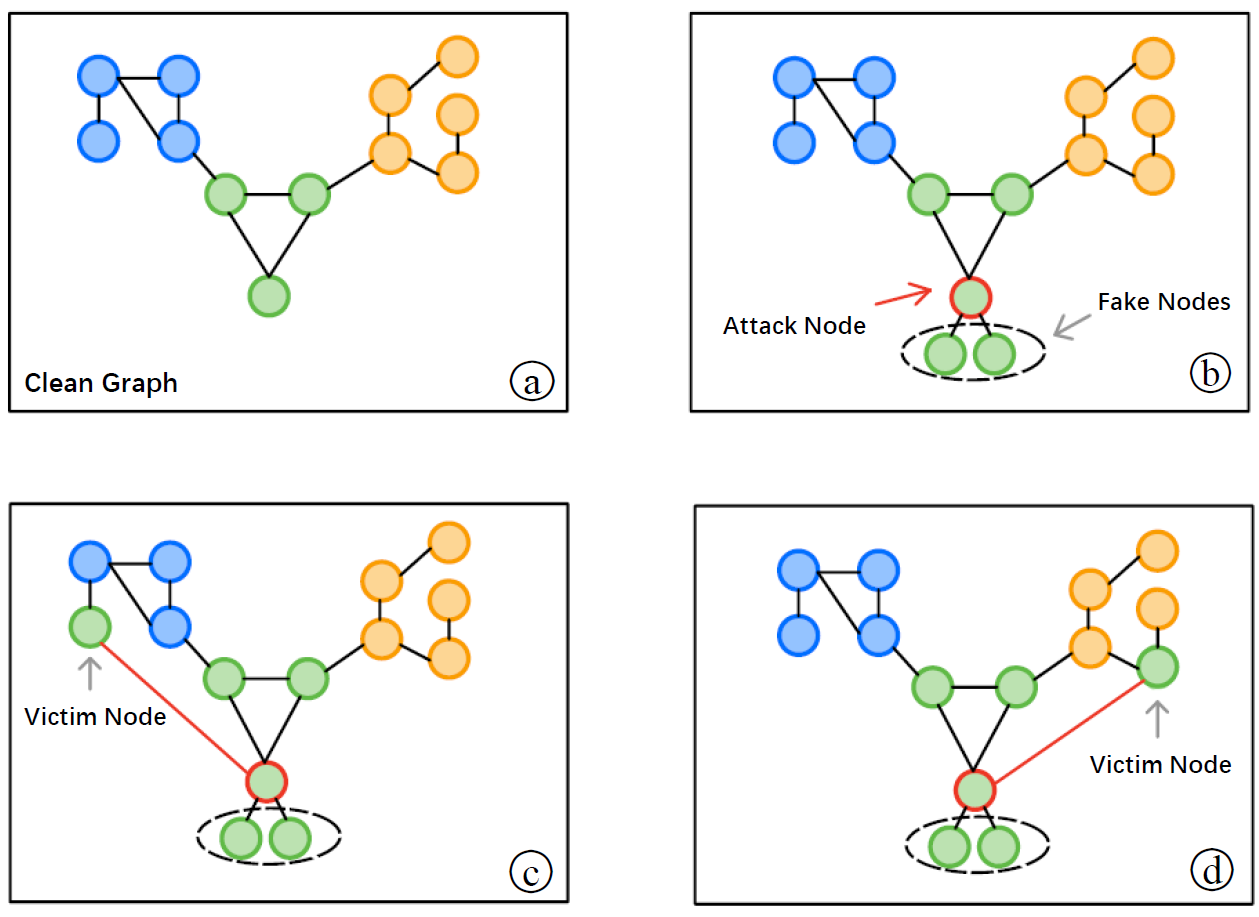}
    \caption{Illustration of the TUA on GCN}
    \label{fig:fig_1}
\end{figure}

In this paper, we propose a targeted universal attack (TUA) on GCN. The TUA is a special universal adversarial attack that can fool GCN into misclassifying any node as the class specified by adversaries. The main idea of the TUA is to pick several nodes in the graph as attack nodes. Then, the attack capability of the attack nodes is enhanced by connecting some fake nodes to them and computing perturbations in the attributes of these fake nodes. After that, when linking to the attack nodes, any node in the graph (called a victim node) will be misclassified by the GCN as the attack node class (called the target class) controlled by the adversary. Figure \ref{fig:fig_1} illustrates the procedures of our attack method, where the nodes in the original graph are divided into three categories, which are represented by green, blue and yellow, respectively (Figure \ref{fig:fig_1}a). We first choose an attack node from the green class and link two fake nodes to it (Figure \ref{fig:fig_1}b). Then, we compute the perturbation in the attributes of the fake nodes to enhance the attack capability of the attack nodes. Finally, if we connect any victim node to the green attack node, the GCN will be fooled to misclassify it as the green class. For example, by linking the blue victim node (Figure \ref{fig:fig_1}c) or yellow victim node (Figure \ref{fig:fig_1}d) to the attack node, those victim nodes will be categorized as the green class.

Given the few attack nodes controlled by the adversary, it is difficult in general for him/her to make the GCN misclassify any node in the graph as the attack node class by simply linking it to the attack nodes. In the TUA proposed in this paper, a few fake nodes with perturbed attributes are connected to the attack nodes to enhance the attack capability of the given attack nodes through the feature aggregation process of the GCN. According to the experimental results on three popular datasets, the average attack success rate of the proposed TUA on any victim node reaches approximately 83\% by employing only 6 fake nodes and 3 attack nodes.

To the best of our knowledge, the graph universal attack (GUA) proposed by Zang et al.\cite{4_TUA} is the only work in this specific field. The GUA is an untargeted attack that aims to fool the GCN to misclassify any victim node as a class other than its ground truth class through edge rewiring between attack nodes and the victim node (rewir-\\ing means that the existing links from the attack nodes to the victim node are removed while nonexisting links are created). In comparison with GUA, the TUA proposed in this paper is a targeted universal adversarial attack, that is, it can fool the GCN to misclassify any victim node as a specified class.

The contributions of our paper are summarized as follows:
\begin{enumerate} [(1)]
    \item We propose a targeted universal adversarial attack against GCNs, which confirms the existence of this new security vulnerability in GCNs. To the best of our knowledge, our research is the first work on targeted universal adversarial attacks on graphs.
    \item The attack method proposed in this paper is easy to implement with a small number of attack nodes controlled by adversary and fake nodes, and a high attack success rate can be achieved through minor topological modifications. 
\end{enumerate}	

The rest of this paper is organized as follows. In section \ref{sec_Related_work}, we introduce the existing researches on adversarial attacks against GCNs. The background of the GCN and the details of the TUA are provided in Section \ref{sec_3_preli} and Section \ref{sec_Attack}, respectively. In Section \ref{sec_Experiment}, we evaluate our attack method on three popular graph datasets. The conclusion and future research direction are presented in Section \ref{sec_Conclusion}.

\section{Related Work}
\label{sec_Related_work}
\subsection{Adversarial attack in graph structured data}
Over the last few years, it has been reported that GCNs are vulnerable to adversarial attacks (Zungner et al.\cite{1_zugner}, Dai et al.\cite{2_Dai}). Based on the different stages of attacks, adversarial attacks in GCNs are categorized into two types: poisoning attacks (attacks during training time) and evasion attacks (attacks during test time). Generally, poisoning attacks focus on decreasing the performance of GCN models by perturbing training data while evasion attacks construct adversarial samples by modifying attributes or topologies so that GCN models are fooled into misclassification. In addition, according to diverse purposes of the attack, adversarial attacks on graph structured data can also be categorized into node classification attacks, link prediction attacks and graph classification attacks. The goal of node classification attacks is to cause certain nodes to be misclassified by a GCN\cite{2_Dai,1_zugner,22_Aleksandar,15_poison_neighbor}. Link prediction attacks focus on decreasing the association between nodes, which will cause the GCN to provide misguided prediction results\cite{22_Aleksandar,23_Chen, 24_Sun}. Graph classification attacks aim to augment the correlation between the specified graph and target class so that GCNs cannot correctly classify the given graph samples\cite{25_Zhang_Zaixi, 26_Tang_Haoteng}. The attack we proposed in this paper can be categorized into evasion attacks and node classification attacks.

Among all the adversarial attacks on graph-structured data, the fake node attack is a common attack method that is implemented by injecting a group of fake nodes into the graph so that topological or attributional modification of the original graph can be avoided. For example, the GreedyAttack and GreedyGAN proposed by Xiaoyun Wang et al.\cite{10_Wang_Xiaoyun} conducted targeted node attacks by adding fake nodes directly to the victim nodes. Jihong Wang et al.\cite{11_Wang_Jihong} introduce the approximate fast gradient sign method, which adds a vicious node between the victim node and other nodes so that the victim node will be misclassified. However, most existing fake node attacks (\cite{11_Wang_Jihong,10_Wang_Xiaoyun,12_Sun_YiWei} ) are not designed to conduct universal adversarial attacks. In the TUA proposed in this paper, the fake nodes act as the 2-hop neighbor of the victim node. Due to the aggression process of the GCN, the impacts of the features of fake nodes are passed through attack nodes to the victim node so as to conduct target universal adversarial attack.

\subsection{Universal Adversarial Attack}
Universal adversarial attacks were first proposed in image classification by Moosavi-Dezfooli et al.\cite{3_universal_att}. Compared to the former adversarial attack methods\cite{8_explaning,9_Intriguing}, universal adversarial attacks are capable of deceiving Convolutional Neural Networks (CNNs) on any sample with the same perturbation. Since the same perturbation can be applied to any sample in the dataset, such an attack method is called a universal adversarial attack.

While universal adversarial attacks have been extensively studied in image classification tasks using CNNs\cite{3_universal_att}, graph universal adversarial attacks have barely been explored. To the best of our knowledge, the GUA proposed by Zang et al.\cite{4_TUA} is the only work in this specific field. The GUA is an untargeted attack that aims at fooling a GCN to misclassify any victim node as a class other than its ground truth through edge rewiring between attack nodes and the victim node. The GUA assumes that the adversary can access any node of the whole graph to choose the attack nodes. Compared with the GUA, the TUA proposed in this paper has the following three differences: First, the TUA is a targeted universal adversarial attack, that is, it can fool the GCN to misclassify any victim node as a specified class. Second, the adversary in the TUA only needs to control a small number of nodes other than all nodes of the whole graph. Moreover, only a few topological modifications of the victim node are necessary to launch a TUA, which makes the TUA much easier to implement.

\section{Preliminaries}
\label{sec_3_preli}

Given an attributed graph with binary features $G\ =\ (\bm{A},\ \bm{X})$, where $\bm{A}\ \in\ \left\{0,1\right\}^{N\times N}$ and $\bm{X}\ \in\left\{0,1\right\}^{N\times d}$ are the adjacency matrix with $N$ nodes and the $d$-dimensional feature matrix of nodes in $G$, respectively, the goal of node classification is to predict the correct label for test nodes by using a small group of labeled training nodes. Let us assume that the nodes in the original graph are $V=\left\{v_1,\ldots,v_N\right\}$, and the prediction will be implemented by learning a function $g:V\rightarrow C$ that projects node $v\ \in V$to a specific class in $C$, where $C=\left\{c^{(1)},c^{(2)},\ldots,c^{(N_c)}\right\}$ is a $N_c$-class label set.While there are multiple versions of GCNs, we will only consider the graph convolutional network introduced by Kipf \& Welling\cite{7_semi_conv} which is a typical transductive learning scenario. Specifically, the GCN aggregates information from neighbors using the following hidden layer:
\begin{equation}\label{eq_1}
    \bm{H}^{\left(l+1\right)}\ =\ \sigma\left({\widetilde{\bm{D}}}^{-\frac{1}{2}}\widetilde{\bm{A}}{\widetilde{\bm{D}}}^{-\frac{1}{2}}\bm{H}^{\left(l\right)}\bm{W}^{\left(l\right)}\right)
\end{equation}
where $\widetilde{\bm{A}}=\bm{A}+\bm{I}_N$ is the adjacency matrix after adding self-loops via the identity matrix $\bm{I}_N$, and ${\widetilde{\bm{D}}}_{ii}=\ \sum_{j}{\widetilde{\bm{A}}}_{ij}$ and $\bm{W}^{\left(l\right)}$ are the degree matrix and trainable weight matrix for the specific layer $l$, respectively. $\sigma\left(x\right)$ denotes an activation function (e.g., ReLU). Beginning with $\bm{H}^{\left(0\right)}=\bm{X}, \bm{H}^{(l)}\in\mathbb{R}^{N\times d}$ is the activated value of the $l^{th}$ layer. According to Kipf \& Welling\cite{7_semi_conv}, we employ a semi-supervised GCN model with a single hidden layer that takes a simple form as follows:
\begin{equation}
\begin{split}
Z &=f\left(\bm{A},\bm{X}\right) \\
&=softmax\left(\hat{\bm{A}}ReLU{\left(\hat{\bm{A}}\bm{X}\bm{W}^{\left(0\right)}\right)\bm{W}^{\left(1\right)}}\right)
\end{split}
\end{equation}
where $\hat{\bm{A}}=\ {\widetilde{\bm{D}}}^{-\frac{1}{2}}\widetilde{\bm{A}}{\widetilde{\bm{D}}}^{-\frac{1}{2}}$ is the symmetrically normalized adjacency matrix.

\section{Attack Algorithms}
\label{sec_Attack}
\subsection{Overview}
We study the TUA on an attributed graph with binary features. The TUA is based on the aggregating characteristic of the GCN: when the GCN aggregates the features of the 1-hop neighbors of one node, the features of its 2-hop neighbors will be passed to it via its 1-hop neighbors. In the TUA scenario, after connecting the victim node to the attack nodes, the attack nodes and fake nodes act as the 1-hop neighbors and 2-hop neighbors of the victim node, respectively. According to the above aggregating characteristic of the GCN, perturbations in the features of fake nodes will be passed to victim nodes through attack nodes, which can lead to the misclassification of the victim nodes by the GCN. We assume that the attack nodes controlled by the adversary in the TUA are all from the same category (called the target class). The TUA consists of following 3 steps. First, we choose a set of attack nodes from the target class and connect a few fake nodes with null features to them, which will be described in Section \ref{subsec_add_fake_nodes}. Second, the perturbation in the features of the fake nodes is iteratively computed based on the gradient of a set of randomly selected nodes (called ancillary nodes), which will be presented in detail in Section \ref{subsec_perturbation}. Finally, any victim node will be misclassified by the GCN as the target class if it is connected to the attack nodes.

\subsection{Adding fake nodes with null features}
\label{subsec_add_fake_nodes}
The goal of our attack is to cause the GCN to misclassify any victim node as the target class $c_A$. A set of fake nodes with null features $V_{Fake}=\{v_{Fake}^{(1)},v_{Fake}^{(2)},\ldots,$ $v_{Fake}^{(N_{Fake})}\}$ are first connected to the attack nodes $V_A=\{v_A^{(1)},v_A^{(2)},\ldots,v_A^{(N_A)}\}$ that belong to the target class $c_A$, where $N_{Fake}$ and $N_A$ represent the numbers of fake nodes and attack nodes, respectively. Specifically, given an attributed graph $G=(\bm{A},\bm{X})$, the attack nodes $V_A$ and the number of fake nodes $N_{Fake}$, we can obtain a new graph $G^\prime\ =\ (\bm{A}^\prime,\bm{X}^\prime)$ from the above $G$ with fake nodes added into it, which is defined as follows: 
\begin{equation}
    \bm{A}^\prime=\ \left[\begin{matrix}\bm{A}&\bm{E}\\\bm{E}^T&\bm{P}\\\end{matrix}\right],\ \bm{X}^\prime=\ \left[\begin{matrix}\bm{X}\\\bm{X}_{Fake}\\\end{matrix}\right]
\end{equation}
where $\bm{E}\in{\{0,1\}}^{N\times N_{Fake}}$ denotes the undirected edges between the attack node and fake nodes, and $\bm{P}\in{\{0,}$ $1\}^{N_{Fake}\times N_{Fake}}$ is initialized to zero. To reduce the complexity of the attack, the same number of fake nodes is linked to each attack node during attack. $\bm{X}_{Fake}\in\{0,$ $1\}^{N_{Fake}\times d}$ is the binary feature matrix of fake nodes, and it is initialized to zero.

\subsection{Computing the perturbation in the features of the fake nodes}
\label{subsec_perturbation}
Given a set of ancillary nodes $V_T=\{{v_T}^{\left(1\right)},{v_T}^{\left(2\right)},\ldots,$ ${v_T}^{\left(N_T\right)}|{v_T}^{\left(:\right)}\in V\cap c_{{v_T}^{\left(:\right)}}\neq c_{v_A}\}$, where any node $v\in V_T$ is not from the target class $c_A$, $N_T$ denotes the number of ancillary nodes and V is the node set of the original graph, the goal of our attack is to compute a perturbation in the features of the fake nodes $\bm{X}_{Fake}$ based on the ancillary node set $V_T$ so that any victim node linked to the attack nodes will be misclassified as the target class under the perturbation. The objective function can be defined as follows:
\begin{equation}\label{eq_4_obj_f}
\begin{split}
    \mathcal{F}\left(\bm{A}^\prime,\bm{X}^\prime,v\right)=& \left[f\left(\bm{A}_{(v,V_A)}^\prime,\bm{X}^\prime\right)\right]_{v,c_A} \\
    & -\left[f\left(\bm{A}_{(v,V_A)}^\prime,\bm{X}^\prime\right)\right]_{v,c_v}
\end{split}
\end{equation}
where $v$ is any ancillary node ($v\in V_T$), and $\bm{A}_{\left(v,V_A\right)}^\prime$ represents the adjacency matrix of $G^\prime$ after connecting ancillary node $v$ to attack nodes $V_A$. $\left[f\left(\cdot\right)\right]_{v,c_A}$ and $\left[f\left(\cdot\right)\right]_{v,c_v}$ denote the output probabilities of the GCN assigning node $v$ to target class $c_A$ and current class $c_v$, respectively. For any ancillary node $v$, the larger the output value of the GCN on a certain class is, the more likely the GCN model classifies node $v$ as this class. In objective function \ref{eq_4_obj_f}, if node $v$ is not classified as target class $c_A$ after being connected to attack nodes $V_A$, the output value of $f\left(\cdot\right)$ on class $c_v$ will be higher than that on class $c_A$, i.e., $\left[f\left(\cdot\right)\right]_{v,c_v}>\left[f\left(\cdot\right)\right]_{v,c_A}$, which results in a negative result of $\mathcal{F}\left(\bm{A}^\prime,\bm{X}^\prime,v_T\right)$. Otherwise, when the GCN categorizes node $v$ as target label $c_A$,$\ \left[f\left(\cdot\right)\right]_{v,c_v}$ will be equal to $\left[f\left(\cdot\right)\right]_{v,c_A}$, so the result of $\mathcal{F}\left(\bm{A}^\prime,\bm{X}^\prime,v\right)$ is 0.

We compute the perturbation in the features of fake nodes with the optimization problem in Eq.(\ref{eq_5_argmax}) so that the output value of $\sum_{v\in V_T}\mathcal{F}\left(\bm{A}^\prime,\bm{X}^\prime,v\right)$ is as large as possible. The larger $\sum_{v\in V_T}\mathcal{F}\left(\bm{A}^\prime,\bm{X}^\prime,v\right)$ is, the more likely any node will be misclassified as the target class when connected to the attack nodes.
\begin{equation}\label{eq_5_argmax}
\begin{split}
        & \argmax_{\bm{X}_{Fake}}{{\sum_{v\in V_T}\mathcal{F}\left(\bm{A}^\prime,\bm{X}^\prime,v\right)}} \\
        & s.t. \left\lVert \bm{E} \right\rVert _0+\left\lVert \bm{X}_{Fake} \right\rVert_0 \leq \Delta
\end{split}
\end{equation}

In Eq.(\ref{eq_5_argmax}), $\left\lVert \bm{E} \right\rVert _0$ represents the number of edges between fake nodes and attack nodes, which is also equivalent to the number of fake nodes injected into the original graph. $\left\lVert \bm{X}_{Fake} \right\rVert_0$ refers to the number of fake node features. Both $\left\lVert \bm{E} \right\rVert _0$ and $\left\lVert \bm{X}_{Fake} \right\rVert_0$ are bounded by a constant $\Delta$.

Inspired by Moosavi-Dezfooli et al.\cite{5_deepfool} who adopted a gradient-based method to compute perturbations to cause CNNs to misclassify images, we propose a gradient-based method to solve the optimization problem in Eq.(\ref{eq_5_argmax}) to compute perturbations in the features of fake nodes. At the beginning of the method, the features of the fake nodes are initialized to 0, i.e., $\bm{X}_{Fake}=0$. Then, we iteratively compute the perturbation in the features of fake nodes. There are two procedures in every iteration as follows:
\begin{algorithm}

    \caption{Compute the perturbations in the features of fake nodes}
    \label{al_1}
    \SetKwInOut{Input}{input}
    \SetKwInOut{Output}{output}

    \Input{Graph $G(\bm{A},\bm{X})$, GCN classifier $\left.f\right.$, the number of fake nodes $N_{Fake}$, max number of iterations $\delta$, attack nodes $V_A$, target label $c_A$, ancillary nodes $V_T$}
    \Output{Perturbated graph $G^\prime=\ \left(\bm{A}^\prime{,\ \bm{X}}^\prime\right)$}
    $\bm{X}^\prime,\bm{A}^\prime\gets$ connect $N_{Fake}$ nodes with attack nodes $V_A$
    
    \For{$iter=\ 1$ to $\delta$ }
    {
    
    Compute $\nabla_{\bm{X}_{Fake}}\mathcal{F}\left(\bm{A}^\prime,\bm{X}^\prime,v_T\right)$ for each ancillary node with Eq.(\ref{eq_4_obj_f})
    
    Compute $\bm{Grad}$ with Eq.(\ref{eq_6_Grad})
    
    $\argmax\limits_{i,j}{\bm{Grad}\left(i,j\right)}$ with Eq.(\ref{eq_7_ij})
    
    Set $\bm{X}_{Fake}(i,j)$ in $\bm{X}_{Fake}$ to 1
    }
    
    \Return $G^\prime(\bm{A}^\prime,\bm{X}^\prime)$
    
\end{algorithm}
\begin{enumerate} [(1)]
    \item 	Locate the feature in $\bm{X}_{Fake}$ that has the greatest influence on increasing the value of $\sum_{v\in V_T}\mathcal{F}\left(\bm{A}^\prime,\bm{X}^\prime,v\right)$ in the current iteration. We can obtain the influences of each feature in the fake nodes on $\sum_{v\in V_T}\mathcal{F}\left(\bm{A}^\prime,\bm{X}^\prime,v\right)$ by computing the partial derivation of $\sum_{v\in V_T}\mathcal{F}(\bm{A}^\prime,$ $\bm{X}^\prime, v)$ w.r.t. every feature in $\bm{X}_{Fake}$, as shown in Eq.(\ref{eq_6_Grad}) and Eq.(\ref{eq_7_ij}).
    \begin{equation}\label{eq_6_Grad}
        \bm{Grad}=\ \sum_{v\in V_T}\left[\nabla_{\bm{X}_{Fake}}\mathcal{F}\left(\bm{A}^\prime,\bm{X}^\prime,v\right)\right]
    \end{equation}
    \begin{equation}\label{eq_7_ij}
    \begin{split}
         & \argmax_{i,j}{\bm{Grad}\left(i,j\right)} \\
         i\ \in1 & \ldots N_{Fake}\ and\ j\in1\ldots d,
    \end{split}
    \end{equation}
    where $d$ is the dimension of the features for each fake node; and $\bm{Grad}$ is a gradient matrix (its dimension is $N_{Fake}\times d$) of $\sum_{v\in V_T}\mathcal{F}\left(\bm{A}^\prime,\bm{X}^\prime,v\right)$ w.r.t. $\bm{X}_{Fake}$, which indicates the influence of the features in $\bm{X}_{Fake}$ on $\sum_{v\in V_T}\mathcal{F}\left(\bm{A}^\prime,\bm{X}^\prime,v\right)$. $\bm{Grad}\left(i,j\right)$ denotes the partial derivation of $\sum_{v\in V_T}\mathcal{F}\left(\bm{A}^\prime,\bm{X}^\prime,v\right)$ with respect to the $j^{th}$ feature of the $i^{th}$ fake node in $\bm{X}_{Fake}$. The larger the value of the element in $\bm{Grad}$ is, the more significant the corresponding feature in $\bm{X}_{Fake}$ affects the target $\sum_{v\in V_T}\mathcal{F}\left(\bm{A}^\prime,\bm{X}^\prime,v\right)$. According to index $(i,j)$ of the maximum element in Grad, we can locate the corresponding fake node feature $\bm{X}_{Fake}(i,j)$, which denotes the $j^{th}$ feature of the $i^{th}$ node in $\bm{X}_{Fake}$ that has the greatest influence on increasing $\sum_{v\in V_T}\mathcal{F}(\bm{A}^\prime,$ $\bm{X}^\prime,v)$ in the current iteration. If the value of $\bm{X}_{Fake}(i,$ $j)$ has already been set to 1, the index $(i,j)$ corresponding to the second largest element in Grad will be used instead, and this repeats.
    \item Set the value of $\bm{X}_{Fake}(i,j)$ to 1, which is the perturbation in the current iteration.
\end{enumerate}

After a few iterations, we can obtain a new graph $G^\prime(\bm{A}^\prime,\bm{X}^\prime)$ where the computed perturbations of $\bm{X}_{Fake}$ in $\bm{X}^\prime$ can enhance the attack capability of attack nodes and lead to the misclassification of any nodes after being connected to the attack nodes. The attack is shown in Algorithm \ref{al_1}.

\section{Experimental Evaluation}
\label{sec_Experiment}
\subsection{Experimental Setup}
In this section, we evaluate the attack success rate of the TUA on graph structured datasets. The attack success rate (ASR) refers to the ratio of successfully attacked samples to the total number of samples. We define the test nodes that are misclassified by the GCN as the successfully attacked samples. The max number of iterations $\delta$ in Algorithm \ref{al_1} is set to 25. The evaluation consists of the following three experiments: (i) test the impact of different numbers of attack nodes on the ASR, (ii) test the impact of different numbers of ancillary nodes on the ASR, and (iii) test the fluctuation of the ASR with different ancillary nodes. The code is available at https://github.com/Nanyuu/TUA.

The experiments are performed using three common attributed graph datasets including Cora (2708 nodes, \\5429 edges, and 1433 features), Citeseer (3312 nodes, 4732 edges, and 3703 features) and Pubmed (19717 nodes, 44338 edges, and 500 features)\cite{27_dataset}. The GCN models are trained on the three respective datasets according to the settings of Kipf \& Welling\cite{7_semi_conv}. Then, we conduct TUAs on these GCN models and evaluate the ASRs.

\subsection{Expediting the Computation of the Perturbations}
Most gradient-based attacks suffer from the problem of high time and memory costs. To settle this issue, Li et al.\cite{13_large_scale} proposed an attack framework for efficient adversarial attacks that attacks a smaller subgraph consisting of the k-hop neighbors of the target node (k depends on the number of GCN layers) so that unnecessary graph information storage and computation can be avoided.\cite{14_SGC,16_calibration}

We implement our TUA based on the above subgraph construction method to improve the computational efficiency in the experiment. Since the nodes are only influenced by their 1-hop and 2-hop neighbors in a 2-layer GCN, we can easily extract a much smaller subgraph centered on the victim nodes or ancillary nodes. Then, the perturbations in the features of fake nodes can be directly computed based on the constructed subgraph, which results in a significant improvement in time efficiency compared to the TUA without a subgraph.

We evaluate the efficiency of our algorithm with and without subgraphs on three datasets, Cora, Citeseer and Pubmed, respectively, according to the number of computations of $\nabla_{\bm{X}_{Fake}}\mathcal{F}\left(\bm{A}^\prime,\bm{X}^\prime,v_T\right)$ per second, which is shown in Table \ref{tabel_1}.

It can be seen from Table \ref{tabel_1} that the TUA is capable of reducing the huge calculation costs using the subgraph construction method. It achieves speedups of 30x, 44x and 59x on Cora, Citeseer and Pubmed, respectively. In the following experiments, we will conduct our attack method with the subgraph construction method to improve the attack efficiency.
\newcommand{\tabincell}[2]{\begin{tabular}{@{}#1@{}}#2\end{tabular}} 
\begin{table}[t]

\caption{The computing efficiency of $\nabla_{\bm{X}_{Fake}}\mathcal{F}\left(\bm{A}^\prime,\bm{X}^\prime,v_T\right)$ with and without the subgraph construction method}

\label{tabel_1}
\begin{tabular}{|l|l|l|}
\hline
Dataset  & \tabincell{c}{Without Subgraph \\(times/s)} & \tabincell{c}{With Subgraph \\(times/s)} \\ \hline
Cora     & 5.12                       & 156.57                  \\ \hline
Citeseer & 4.29                       & 189.28                  \\ \hline
Pubmed   & 1.21                       & 71.38                   \\ \hline
\end{tabular}
\end{table}

\subsection{Experimental Results and Discussion}
\subsubsection{ Influence of the Number of Attack nodes on ASR}

\begin{figure*}[htbp]
    \centering
    \begin{minipage}[t]{0.48\textwidth}
        \centering
        \includegraphics[scale=0.85]{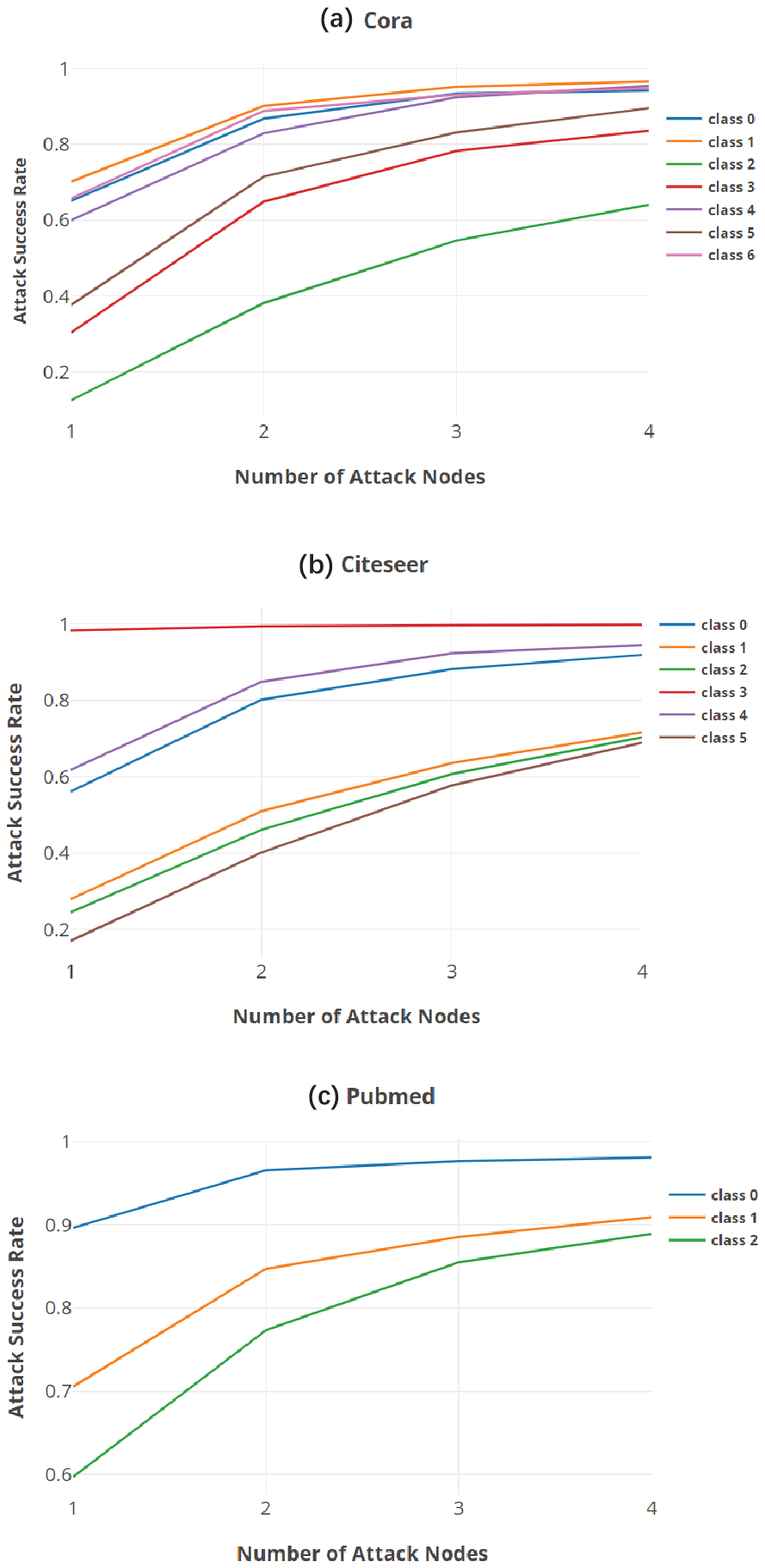}
        \caption{The change of the average ASR with the number \\ of attack nodes.}
        \label{Fig_2}
    \end{minipage}
    \begin{minipage}[t]{0.48\textwidth}
        \centering
        \includegraphics[scale=0.85]{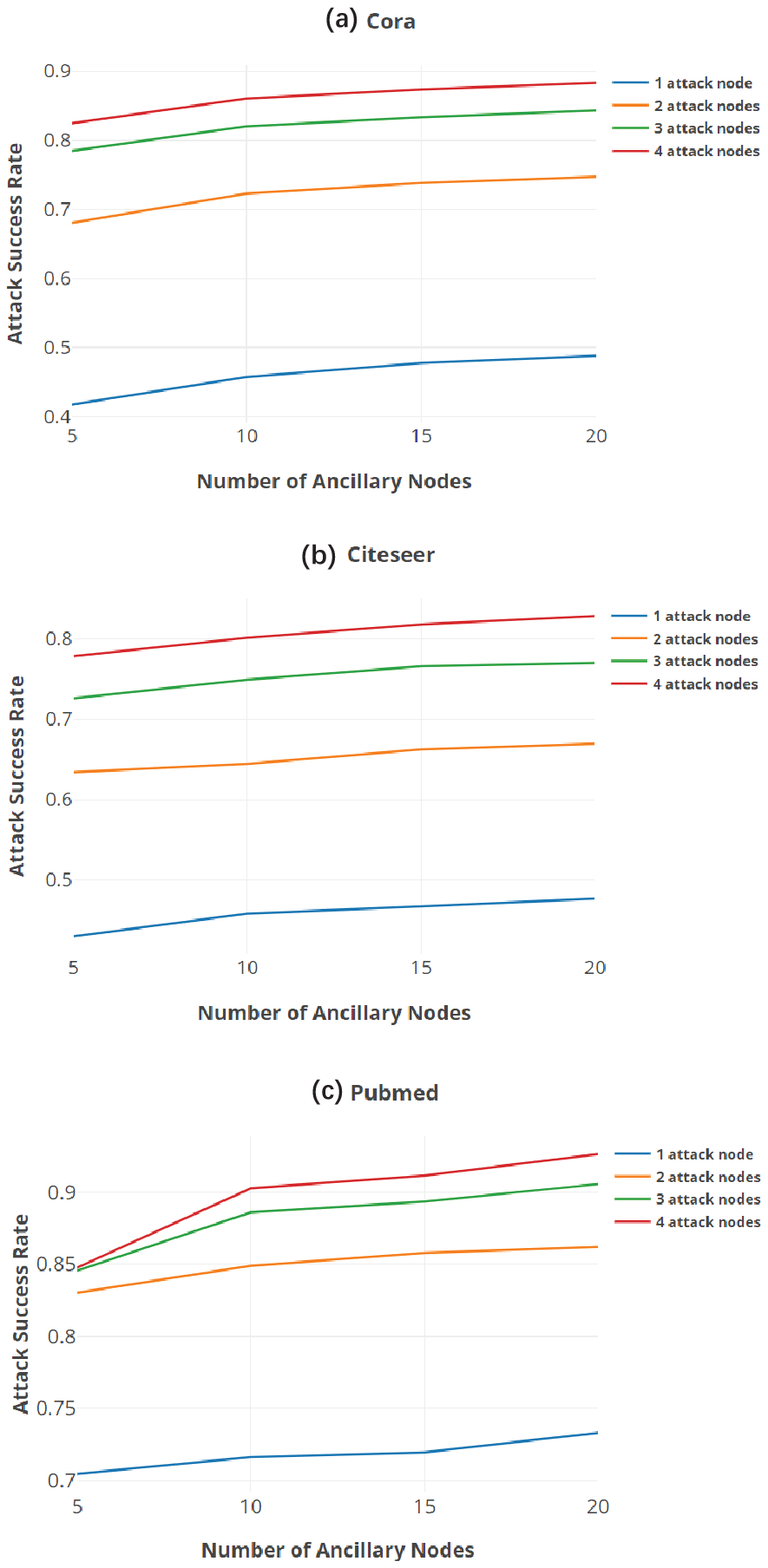}
        \caption{The change of the average ASR with the number \\ of ancillary nodes.}
        \label{Fig_3}
    \end{minipage}
\end{figure*}

\begin{figure*}
    \centering
    \includegraphics[scale=0.2]{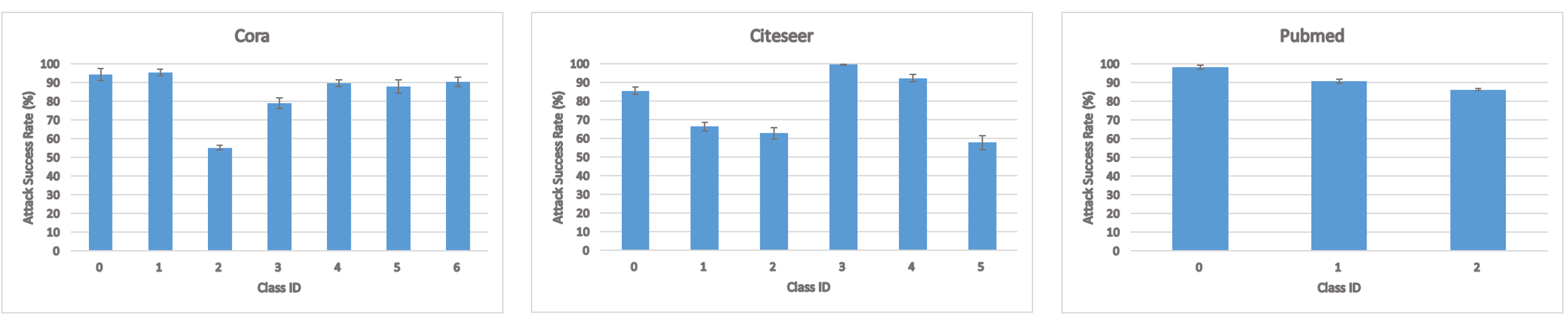}
    \caption{Fluctuation of the ASR with randomly selected ancillary nodes.}
    \label{fig_4_STD}
\end{figure*}
    
    This section evaluates the impact of different numbers of attack nodes on the ASR of the TUA. First, we create sets of different numbers of attack nodes (i.e., 1, 2, 3 and 4, respectively) that are randomly selected from every class of the Cora, Citeseer and Pubmed datasets, respectively. Then, for each of these sets, 2 fake nodes with null features are connected to each of the attack nodes, and perturbations in the features of fake nodes are computed based on 20 ancillary nodes that are randomly picked from the graph. Finally, other nodes in the graph that exclude the above ancillary nodes and those belonging to the target class are selected as test nodes to evaluate the ASR. 
    
    The total number of experiments we conduct is 640. The experimental results with respect to the different numbers of attack nodes are listed in Table \ref{tabel_2}, where the ‘Class ID’ in every row represents the classes of the above three datasets from which attack nodes are selected, and every percentage in the same row is the average ASR corresponding to different numbers of attack nodes. For example, in the third row of the table, the dataset is ‘Cora’; the class ID is 0; the number of attack nodes selected from class ID 0 are 1, 2, 3 and 4, respectively; and the corresponding average ASRs with these attack nodes are 65.1\%, 86.8\%, 93.4\% and 94.1\%, respectively.

    \begin{table}[t]
    \caption{The average ASR corresponding to different numbers of attack nodes.}
    \label{tabel_2}
\begin{tabular}{|l|l|l|l|l|l|}
\hline
\multirow{2}{*}{Dataset}  & \multirow{2}{*}{Class   ID} & \multicolumn{4}{l|}{The Number of Attack Nodes} \\ \cline{3-6} 
                          &                             & 1          & 2          & 3         & 4         \\ \hline
\multirow{7}{*}{Cora}     & 0                           & 65.1\%     & 86.8\%     & 93.4\%    & 94.1\%    \\ \cline{2-6} 
                          & 1                           & 70.2\%     & 90.1\%     & 95.1\%    & 96.5\%    \\ \cline{2-6} 
                          & 2                           & 12.5\%     & 38.2\%     & 54.6\%    & 64.0\%    \\ \cline{2-6} 
                          & 3                           & 30.4\%     & 64.9\%     & 78.3\%    & 83.6\%    \\ \cline{2-6} 
                          & 4                           & 60.0\%     & 82.9\%     & 92.4\%    & 95.4\%    \\ \cline{2-6} 
                          & 5                           & 37.7\%     & 71.5\%     & 83.1\%    & 89.5\%    \\ \cline{2-6} 
                          & 6                           & 65.7\%     & 88.8\%     & 93.1\%    & 94.8\%    \\ \hline
\multirow{6}{*}{Citeseer} & 0                           & 56.2\%     & 80.2\%     & 88.1\%    & 91.9\%    \\ \cline{2-6} 
                          & 1                           & 28.0\%     & 51.0\%     & 63.6\%    & 71.6\%    \\ \cline{2-6} 
                          & 2                           & 24.6\%     & 46.1\%     & 60.7\%    & 70.3\%    \\ \cline{2-6} 
                          & 3                           & 98.3\%     & 99.4\%     & 99.6\%    & 99.8\%    \\ \cline{2-6} 
                          & 4                           & 61.8\%     & 84.7\%     & 92.3\%    & 94.5\%    \\ \cline{2-6} 
                          & 5                           & 17.1\%     & 40.2\%     & 57.7\%    & 68.9\%    \\ \hline
\multirow{3}{*}{Pubmed}   & 0                           & 89.6\%     & 96.6\%     & 97.6\%    & 98.1\%    \\ \cline{2-6} 
                          & 1                           & 70.6\%     & 84.7\%     & 88.5\%    & 90.9\%    \\ \cline{2-6} 
                          & 2                           & 59.7\%     & 77.3\%     & 85.5\%    & 88.9\%    \\ \hline
\end{tabular}
    \end{table}
    
    It can be seen from Table \ref{tabel_2} that the average ASR reaches 83\% with only three attack nodes. The diagrams in Figure \ref{Fig_2} visualize Table \ref{tabel_2} and use different color lines to demonstrate how the average ASR changes with the number of attack nodes from every class in the three datasets. It can be seen in Figure \ref{Fig_2} that all of the average ASRs rise as the number of attack nodes increases, but most of their rising tends to slow down when the number of attack nodes is greater than 3.
    
   \subsubsection{Influence of the Number of Ancillary nodes on ASR}
    
    This section evaluates how the number of ancillary nodes influences the ASR. First, we create sets of different numbers of attack nodes (i.e., 1, 2, 3 and 4, respectively) that are randomly selected from every class of the Cora, Citeseer and Pubmed datasets, respectively. Then, for each of these sets, two fake nodes are linked to every attack node, and 5, 10, 15 and 20 ancillary nodes that exclude the nodes from the target class are randomly selected from the same graph to compute the perturbations in the features of the fake nodes. Finally, other nodes in the graph that exclude the above ancillary nodes and those belonging to the target class are selected as test nodes to evaluate the ASR. We conduct this experiment 2560 times according to each of the above settings, and the average ASRs are shown in Figure \ref{Fig_3}.
    
    The diagrams in Figure \ref{Fig_3} demonstrate the change in the ASR with the number of ancillary nodes from the three respective datasets using lines with different colors corresponding to different numbers of attack nodes. It can be seen from Figure \ref{Fig_3} that the ASR rises as the number of ancillary nodes increases. However, the change in the ASR tends to gradually slow down when the number of ancillary nodes is greater than 15, which means that it is not necessary to launch a TUA with too many ancillary nodes.
    
    \subsubsection{Fluctuation of ASR with the randomly selected ancillary nodes}
    
    In this section, we test the fluctuation of the ASR with randomly selected ancillary nodes from the Cora, Citeseer and Pubmed datasets, respectively. According to previous experiments, we can achieve a high ASR with 20 ancillary nodes and 3 attack nodes, each of which connects 2 fake nodes. Therefore, we adopt the same setting in this experiment. First, we create sets with 3 attack nodes that are randomly selected from every class of the Cora, Citeseer and Pubmed datasets, respectively; and each of the attack nodes in the set is linked with two fake nodes. Then, for every set of attack nodes mentioned above, we compute the perturbations in the features of the fake nodes 10 times with 20 randomly selected ancillary nodes each time. Finally, for every perturbation, the other nodes in the graph that exclude the above ancillary nodes and those belonging to the target class are selected as test nodes to evaluate the ASR and its standard deviation. The experimental results are shown in Table \ref{tabel_3}, where each row is the average ASR and its standard deviations.
    
    \begin{table}[]
    \caption{The fluctuation of the ASR with randomly selected ancillary nodes}
    \label{tabel_3}
        \begin{tabular}{|l|l|l|l|}
        \hline
        Dataset                   & Class ID & \tabincell{c}{Average  \\ ASR (\%)} & \tabincell{c}{Standard \\Deviation (\%)} \\ \hline
        \multirow{7}{*}{Cora}     & 0        & 94.4               & 3.1                     \\ \cline{2-4} 
                                  & 1        & 95.5               & 1.9                     \\ \cline{2-4} 
                                  & 2        & 55.3               & 1.1                     \\ \cline{2-4} 
                                  & 3        & 79.0               & 2.9                     \\ \cline{2-4} 
                                  & 4        & 89.7               & 1.8                     \\ \cline{2-4} 
                                  & 5        & 88.0               & 3.5                     \\ \cline{2-4} 
                                  & 6        & 90.4               & 2.4                     \\ \hline
        \multirow{6}{*}{Citeseer} & 0        & 85.6               & 1.9                     \\ \cline{2-4} 
                                  & 1        & 66.4               & 2.3                     \\ \cline{2-4} 
                                  & 2        & 62.9               & 3.0                     \\ \cline{2-4} 
                                  & 3        & 99.6               & 0.1                     \\ \cline{2-4} 
                                  & 4        & 92.4               & 2.1                     \\ \cline{2-4} 
                                  & 5        & 57.9               & 3.8                     \\ \hline
        \multirow{3}{*}{Pubmed}   & 0        & 98.3               & 1.1                     \\ \cline{2-4} 
                                  & 1        & 90.8               & 1.1                     \\ \cline{2-4} 
                                  & 2        & 86.3               & 0.5                     \\ \hline
        \end{tabular}
    \end{table}
    
    Figure \ref{fig_4_STD} is the visualization of Table \ref{tabel_3}, where the x-axis represents the classes in three datasets from which the three attack nodes are selected, the y-axis represents the corresponding average ASR, and the error bars show the standard deviation of the average ASR. From Table \ref{tabel_3} and Figure \ref{fig_4_STD}, it can be seen that the average ASR is stable under randomly selected ancillary nodes.
    
\section{Conclusion}
\label{sec_Conclusion}
The GCN is a widely used graph neural network. However, there are few researches studying universal adversarial attacks on GCNs. In this paper, we proposed a targeted universal attack against GCNs. By connecting a small number of fake nodes to the attack nodes and iteratively computing perturbations in the features of the fake nodes w.r.t. the ancillary nodes. The proposed attack can achieve an average attack success rate of 83\% on any node of the original graph under 3 attack nodes and 6 fake nodes. In the future, we will further study the transferability of our universal adversarial attack on different GCN models and the defense methods against it.

\section{Acknowledgement}
This work was supported by the State Scholarship Fund of the China Scholarship Council (Grant No.201606895018).

\bibliography{mybibfile}

\end{document}